\documentclass[letterpaper, 10 pt, conference]{ieeeconf}  

\IEEEoverridecommandlockouts                              
\usepackage{mathpazo} 

\usepackage{authblk}
\usepackage[utf8]{inputenc}
\usepackage{tikz}
\usepackage{pgfplots}
\usepackage{graphicx} 
\usepackage{pgf-umlsd}
\usepackage{mathtools, xparse}
\let\labelindent\relax
\usepackage{enumitem}
\usepackage{todonotes}
\usepackage{amsmath}
\usepackage[T1]{fontenc}
\usepackage[english]{babel}

\title{\LARGE \bf
Segmentation of Surgical Instruments\\
for Minimally-Invasive Robot-Assisted Procedures\\
Using Generative Deep Neural Networks
}

\author[1]{Iñigo Azqueta-Gavaldon}
\author[1]{Florian Fröhlich}
\author[1]{Klaus Strobl}
\author[1]{Rudolph Triebel}
\affil[1]{Institute of Robotics and Mechatronics\\
    German Aerospace Center \protect\\
    Münchener Str. 20, 82234 Weßling, Germany
}

\pgfplotsset{compat=1.14}
\begin{document}

\maketitle
\thispagestyle{empty}
\pagestyle{empty}

\begin{abstract}

This work proves that semantic segmentation on minimally invasive surgical instruments can be improved by using training data that has been augmented through domain adaptation. The benefit of this method is twofold. Firstly, it suppresses the need of manually labeling thousands of images by transforming synthetic data into realistic-looking data. To achieve this, a CycleGAN model is used, which transforms a source dataset to approximate the domain distribution of a target dataset. Secondly, this newly generated data with perfect labels is utilized to train a semantic segmentation neural network, U-Net.  This method shows generalization capabilities on data with variability regarding its rotation- position- and lighting conditions.  Nevertheless, one of the caveats of this approach is that the model is unable to generalize well to other surgical instruments with a different shape from the one used for training. This is driven by the lack of a high variance in the geometric distribution of the training data. Future work will focus on making the model more scale-invariant and able to adapt to other types of surgical instruments previously unseen by the training.

\end{abstract}

\section{INTRODUCTION}

Robot-assisted surgery has experienced an increasing level of acceptance in the medical community in the last years, which together with technological innovation and improvements in robotics may render these systems ubiquitous in the near future. Likewise, there has been a paradigm shift in surgical procedures with the increase of minimally invasive surgeries, which greatly reduce the pain, hospitalization time, and post-surgery complications of the patient. These two developments go hand in hand, since robotic systems can help overcome some of the limitations of traditional laparoscopic Minimally Invasive Surgery (MIS).\\

\begin{figure} [ht]
\includegraphics[height=50mm]{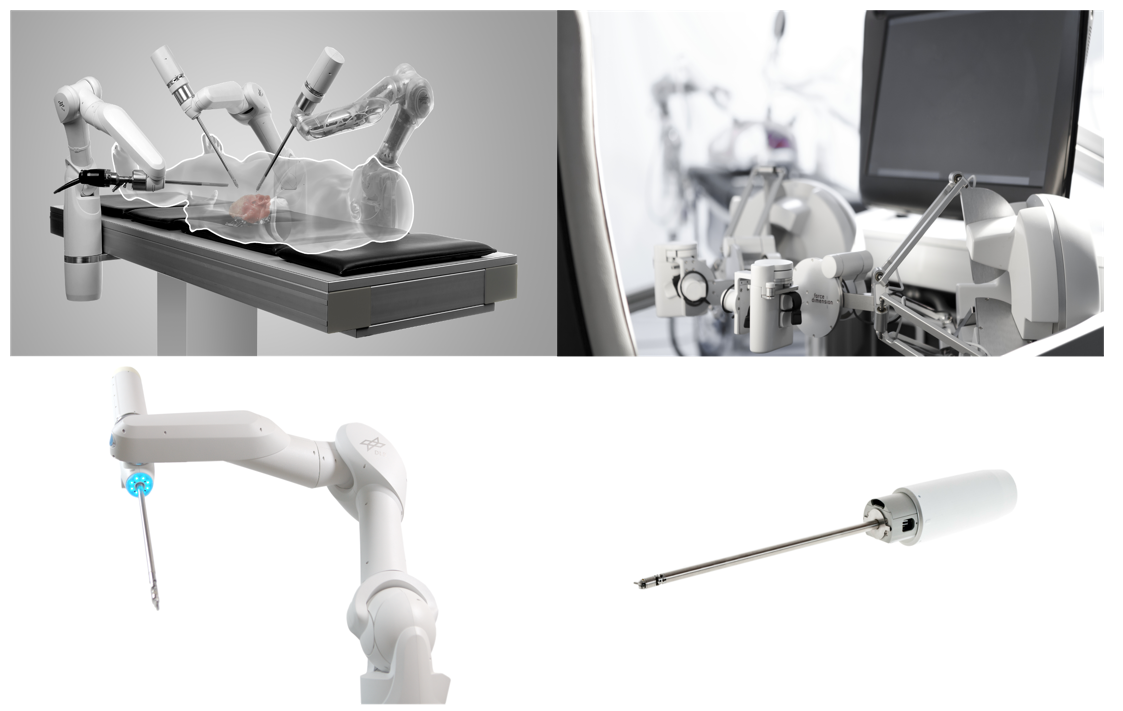}
\centering
\caption{MiroSurge system}
\label{fig:MiroSurge}
\small
 Top left: MiroSurge system with its three robotic arms. \break 
 Top right: User interface and robot arm controlers. \break 
 Bottom left: MIRO robotic arm with MICA instrument. \break Bottom right: Close up of MICA instrument and driving unit.  
\end{figure}

The Institute of robotics and mechatronics of the German Aerospace Center (DLR) has been developing for over a decade a robot-assisted surgical system: the MiroSurge \cite{hagn2008dlr}.  This system consists of three arms with seven degrees of freedom (DoF) that can be teleoperated by a bi-manual control interface.  Two of the arms (MIRO arms) can be equipped with a wide range of instruments, while the third one is equipped with a stereo vision endoscope.  One of the most widely used types of instruments attached to the system is the MICA tool.  It consists of a drive unit for a tool or end-effector, which depending on the task can have up to three DoF. Figure \ref{fig:MiroSurge} shows the MiroSurge, MICA and MIRO setup. \\

Surgeons using a teleoperated system for a minimally invasive surgery suffer two main disadvantages with respect to an open surgery scenario. Their field of view is greatly reduced and the current robotic systems do not offer haptic feed-back, which surgeons rely on in the traditional procedure. These robotic systems can, however, aid the surgeons increase their performance by reducing physiologic tremor, scaling motion, and increasing manual dexterity.  Additionally, algorithms can be developed to further aid the surgeon during operations, for instance by preventing certain tool movements that could cause damage to organs, vessels, etc.\\

The development of such complex algorithms requires the exact knowledge of the pose (position and orientation relative to a reference system) of the surgical tools, in order to track them. Its extraction through forward kinematics is not sufficiently accurate due to intrinsic hardware limitations. On the other hand, computer vision approaches such as object detection and tracking, registration, reconstruction, etc., show a better potential to solving this problem. A key step for the development of such computer vision tools is the ability to distinguish the foreground from the background in the endoscopic image, that is, to be able to differentiate what pixels on the image belong to the surgical instrument (foreground) and what pixels to organs, tissue, etc., (background). This is known as semantic segmentation.\\

To this date, semantic segmentation is best achieved by using deep learning methods in a supervised manner (i.e., leveraging a large labeled dataset). A major challenge of this approach is the size of the datasets needed. Gathering and labeling accurately datasets is a highly time and resource consuming task. Synthetic rendering engines can alleviate this problem by generating a virtually unlimited number of labeled samples, with perfect segmentation labels and even instrument poses. However, these synthetic data do not always capture all the characteristics and features inherent of actual endoscopic images. As a consequence, models trained with rendered synthetic data are prone to overfitting to features only present in the synthetic data, while neglecting important features of the real-world data. In this paper, we propose the use of domain adaptation techniques to create a realistic-looking dataset of MiroSurge surgical instruments to train a semantic segmentation network.  \\

\textbf{Contributions}

Our method describes a two-step approach that combines domain adaptation and semantic segmentation models. To the best of our knowledge, this is the first time a semantic segmentation problem of robot assisted surgical instruments is solved by training a segmentation network with domain-adapted synthetic data.\\

The approach to tackle this problem is influenced by two intrinsic limitations of the MiroSurge system: 
\begin{enumerate}
    \item It is not possible to capture endoscopic and synthetic images with an exact pixel-to-pixel correspondence, due to intrinsic limitations of the system.
    \item Due to time constraints, the size of the dataset is no larger than a few thousand images.
\end{enumerate}

A CycleGAN model \cite{DBLP:journals/corr/ZhuPIE17} was thus chosen to perform the domain adaptation task, due to its capability to be trained with unpaired datasets.\\
A U-Net model \cite{DBLP:journals/corr/RonnebergerFB15} was chosen for the semantic segmentation problem due to its state of the art performance in medical imaging segmentation, especially with a low number of training images (hundreds or a few thousand).

\section{STATE OF THE ART}

The use of domain adaptation methods for the creation of augmented datasets has been proposed in several research papers. Generative Adversarial Networks, introduced in 2014 by Goodfellow \textit{et al.} \cite{goodfellow2014generative} proved to achieve state of the art results in generating realistic-looking images for a specific target domain. For this reason, GANs have been used as the backbone of much of the research focused on domain adaptation or style transfer. 
Shrivastava \textit{et al.} \cite{DBLP:journals/corr/ShrivastavaPTSW16} developed a model named SimGAN to transform computer generated images of eyes and hands into realistic-looking images, while maintaining the input annotations such as gaze direction of the eyes and hand position. They then used the domain-adapted dataset to train gaze and hand position estimation models. Similarly, Wild \textit{et al.} \cite{DBLP:journals/corr/SixtWL16} created another generative model based on GANs named RenderGAN. With this model, they augmented synthetic renderings of markers placed on honeybees, transforming them into more realistic-looking images while also maintaining the input annotations.\\

Semantic segmentation of surgical instruments is a very complex problem due to lighting conditions, occlusions, shadows, glimmer, reflections, etc. Shvets \textit{et al.} \cite{DBLP:journals/corr/abs-1803-01207} used a U-Net architecture \cite{DBLP:journals/corr/RonnebergerFB15} to semantically segment images of a robot assisted surgery. They modified the U-Net architecture by substituting the first contracting half of the network (encoder) by a pre-trained model. The VGG \cite{simonyan2014very} and ResNet \cite{he2016deep} models were both tested, which resulted in an improvement of the segmentation results, from 75$\%$ to 83$\%$ intersection over union, or IoU.
Similarly, Pakhomov \textit{et al.} \cite{DBLP:journals/corr/PakhomovPAAN17} proposed the use of a deep residual network to tackle the same problem on da Vinci surgical instruments. By modifying the ResNet architecture and keeping it fully convolutional instead of fully connected, they were able to use it for semantic segmentation. Additionally, they expanded the scope of the segmentation problem by performing multi-class segmentation, reaching results of up to 83$\%$ IoU.\\

Hanisch \textit{et al.} \cite{hanischsegmentierung} proposed the use of a U-Net architecture to perform semantic segmentation on the surgical instruments of the MiroSurge system. With their approach, the authors were able to achieve semantic segmentation results of up to 77.76\% IoU.

\section{METHOD}

\begin{figure} [ht]
\includegraphics[height=90mm]{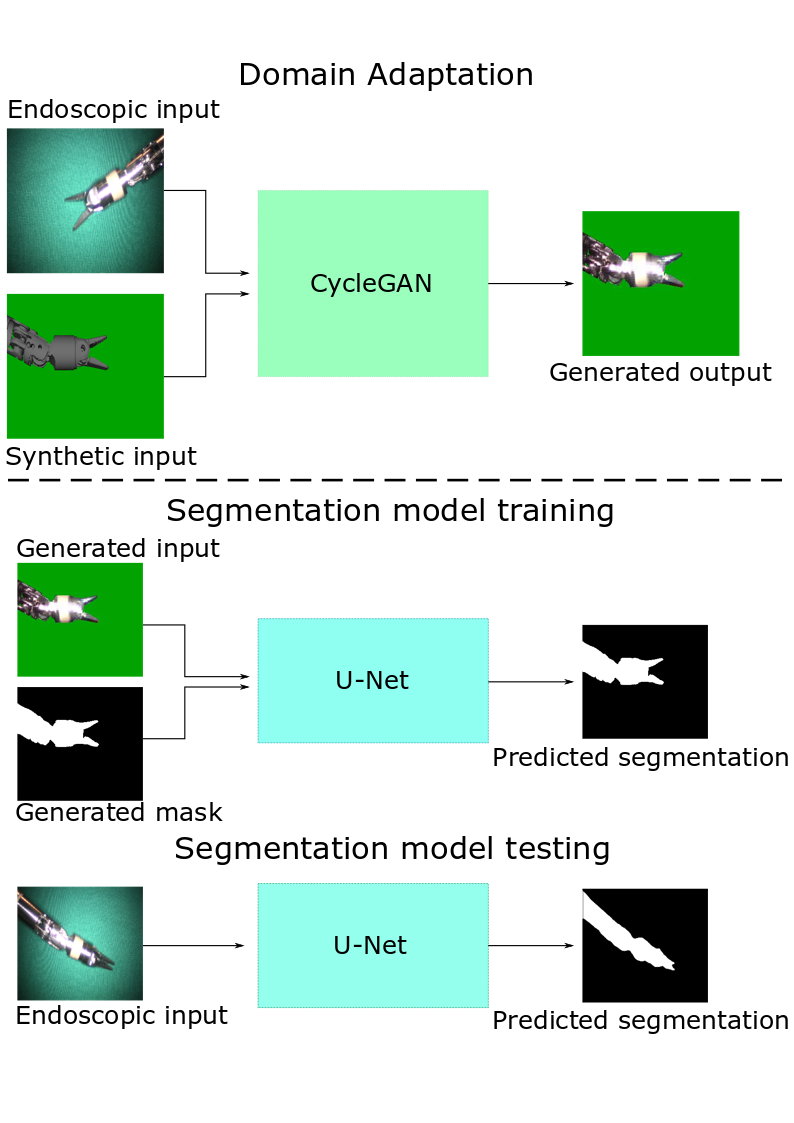}
\centering
\caption{Pipeline of proposed method overview}
\label{Pipeline OVerview}
\footnotesize
\end{figure}

The pipeline of the proposed model can be better visualized in figure \ref{Pipeline OVerview}. First, a CycleGAN model is trained to generate realistic-looking endoscopic images from synthetic data. Then, the domain-adapted generated dataset is used to train a U-Net model to perform semantic segmentation.

\subsection{DOMAIN ADAPTATION}

\textbf{Dataset Creation}\\
In order to perform domain adaptation on synthetic images, two datasets were created. One containing endoscopic images of the surgical instrument against a green-blue background, and another dataset comprised of synthetic images of the surgical instrument from the CAD model, rendered with the in-house software MediView. During their creation, special care was given to two aspects:
\begin{enumerate}
    \item The variance of the pose, field of view, joint angles, and scale of the surgical instruments should be as high as possible. This ensures that the training dataset will yield a segmentation network with better generalization capabilities. Furthermore, the aforementioned scene characteristics are harder to implement as augmentations, whereas lighting conditions, gamma correction, blurring, noise, etc., can all easily be augmented during training.
    \item The pose, field of view, joint angles, and scale of the surgical instruments should be as close as possible in both the endoscopic and synthetic images. This helps preventing the domain adaptation model from finding such differences and focusing on them to transfer from one style to another, instead of focusing on the overall pixel distribution appearance. To achieve this, the actual forward kinematics information from the MiroSurge surgical system was used to render a CAD model of the system. The endoscopic images and snapshots of the CAD model were then taken \textit{simultaneously}.
\end{enumerate}

\begin{figure} [ht]
\includegraphics[height=32mm]{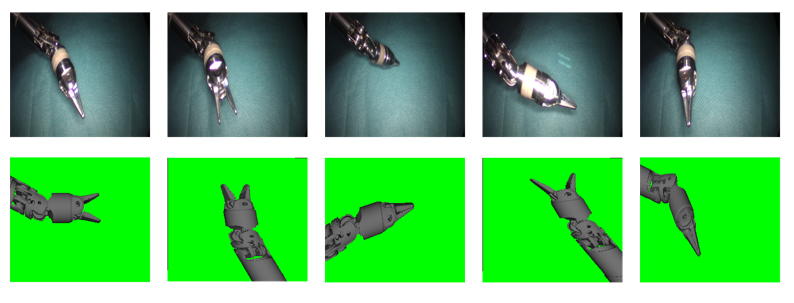}
\centering
\caption{Datasets for domain adaptation}
\label{Datasets for Domain Adaptation}
\footnotesize
Top Row: Real-world endoscopic images.\break Bottom Row: Synthetic rendered images
\end{figure}

Figure \ref{Datasets for Domain Adaptation} shows a sample of both datasets.

\textbf{Training}\\
A CycleGAN network \cite{DBLP:journals/corr/ZhuPIE17} was used as the backbone for the domain adaptation task. Following the approach of Heusel \textit{et al.} \cite{heusel2017gans}, different learning rates were tested for the generators and discriminators, and it was found that a discriminator learning rate ten times larger than the one from the generators yielded the best results. Additionally, a rough manual segmentation of the real endoscopic training images was done to be able to easily introduce random backgrounds on the scene. The green-blue background was not sufficiently uniform to enable an automatic segmentation of sufficient quality. Furthermore, failing to randomize and increase the variance of the backgrounds resulted on the CycleGAN model exclusively trying to approximate the pixel distribution of the green-blue backgrounds, rather than that of the surgical instruments.\\

On table \ref{tab:CycleGAN_training1} the hyperparemeters used to train the CycleGAN that yielded the best results are summarized.\\

\begin{table}[ht]
\caption{Hyperparameters of CycleGAN model}
\label{tab:CycleGAN_training1}
\centering
\begin{tabular}{l l}
Hyperparameter & Value\\

  \noalign{\global\arrayrulewidth=0.4mm}
\hline
Training Images & 2000\\ 
  \noalign{\global\arrayrulewidth=0.1mm}
\hline
Iterations & 30000 \\ \hline
Normalization & Instance Normalization \\ \hline
Batch Size & 1 \\ \hline
Optimizer & Adam \\ \hline
Generator Learning Rate & $2e - 4$  \\ \hline
Discriminator Learning Rate & $2e - 5$  \\ \hline

Adam Optimizer Momentum $\beta$ & 0.5 \\ \hline
Image Pool size & 50 \\ \hline
Cycle Loss Weight $\lambda$ & 10 \\ \hline

\end{tabular}
\end{table}

\begin{figure} [ht]
\includegraphics[height=50mm]{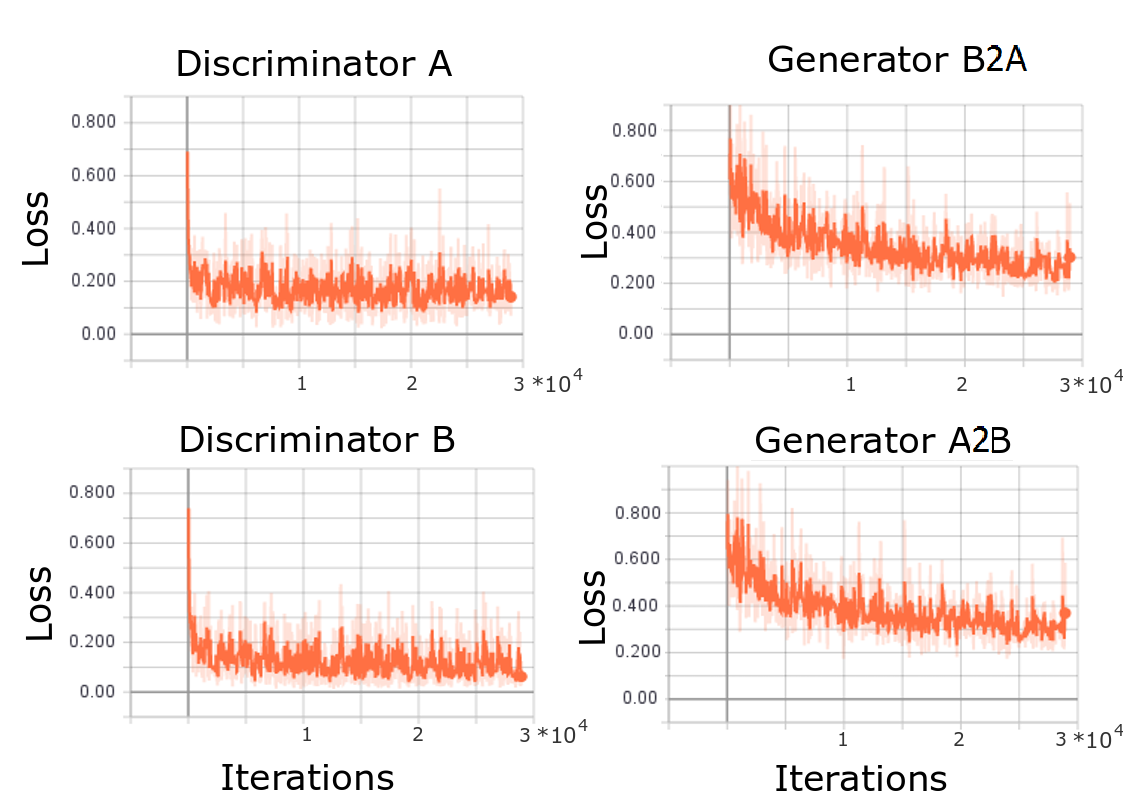}
\centering
\caption{Generator and discriminator losses during CycleGAN training}
\label{CycleGAN_DiscLR_10x}
\end{figure}

With these parameters, the CycleGAN model was trained until the generators of the model converged to a local minimum as shown in figure \ref{CycleGAN_DiscLR_10x}. 

Discriminator A refers to the network that classifies between real or generated endoscopic images. Discriminator B determines whether a given synthetic images is real or has been generated. On the other hand, Generator B2A transforms endoscopic images into synthetic images, and A2B does the opposite, transforming synthetic into endoscopic images.\\

\begin{figure} [ht]
\includegraphics[height=32mm]{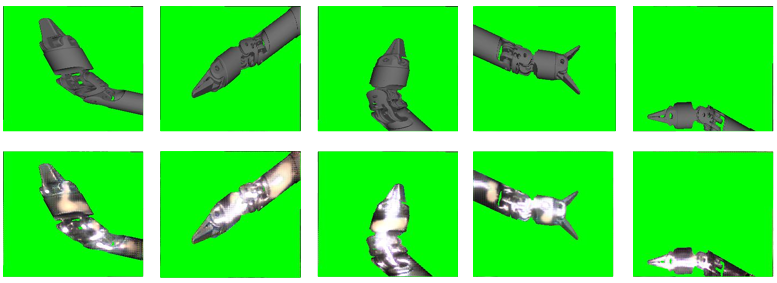}
\centering
\caption{CycleGAN generated examples}
\label{CycleGAN Generated Examples 1}
\end{figure}

At this point, the model yielded subjectively satisfactory results as the ones shown in figure \ref{CycleGAN Generated Examples 1}.

\subsection{SEMANTIC SEGMENTATION}
A U-Net model \cite{DBLP:journals/corr/RonnebergerFB15} was the model of choice for this work due to the state of the art results demonstrated on medical images segmentation problems.
A training dataset of $2500$ domain-adapted realistic-looking images was generated as explained above. Since the domain-transferred images belonged originally to the synthetic dataset, and the content of the images was not changed, perfect segmentation labels were available.\\

\textbf{Augmentations}\\
In order for the network to learn from data with a high variance, the training images were subjected to several augmentations before training. This is common practice in machine learning. By performing augmentations on the training dataset, the probability that the pixel distribution of a previously unseen image lies within the pixel distribution of the learned data is increased. The generalization capabilities of the model are improved in this manner. \\

The available Python library ImgAug was used \cite{ImgAugCitation}. This versatile library offers a comprehensive set of augmentations used in computer vision tasks such as noise filters, occlusion generators, affine transformations, etc. The augmentations performed were the following:

\begin{itemize}
    \item Color channel inversion
    \item Addition of random values between -50 and 50 to image channels
    \item Image multiplication of random values between 0.25 and 1.5
    \item Contrast normalization
    \item Addition of salt and pepper noise
    \item Edge blurring
    \item Addition of random black squared patches
    \item Elastic transformation
\end{itemize}

Additionally a percentage of the total number of images where each augmentation is applied must be defined.\\

\textbf{Training}\\
The U-net architecture was kept as in the original paper \cite{DBLP:journals/corr/RonnebergerFB15} in all the experiments. In their work, the authors used an Adam optimizer, which is a widely used optimization method \cite{kingma2014adam}. They proposed the use of a high momentum $\beta = 0.99$, for the weight update to take into account a large number of previously seen examples. In this way, the authors were able to feed into the network large image tiles and a mini-batch of a single image, since they were restricted by the nature of their training images, which have a sparse distribution of objects of interest. The dataset used in our work does not suffer from that restriction, since for every image there is always a surgical instrument (object of interest) of roughly constant proportions. For this reason, several experiments were performed with different momentum values $\beta$ of the Adam optimizer and different batch sizes.

\section{RESULTS}
\subsection{SAME TOOL}

To estimate the generalization capabilities of the U-Net trained with the generated data, three datasets of 200 images were used to test the model. The first test set is composed of the endoscopic images of the instruments captured against a green-blue background (GrBG). The second test set consists of real endoscopic images with an added background of different organ images (OrBG), and the third has endoscopic images with added organ backgrounds and heavy augmentations (AuBG).\\

\begin{figure} [ht]
\includegraphics[height=40mm]{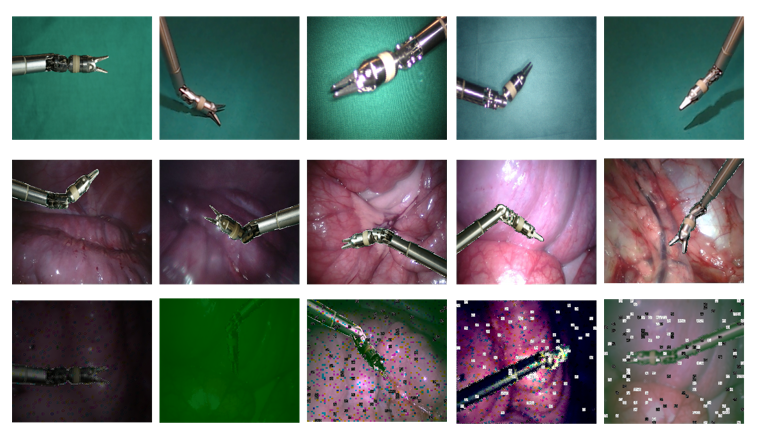}
\centering
\caption{Overview of test datasets}
\label{Test Datasets 1 2 3 Overview}
\footnotesize
Top: GrBG test set. Endoscopic images against green-blue background.\break
Middle: OrBG test set. Endoscopic images with added organ backgrounds.\break 
Bottom: AuBG test set. Endoscopic images with added organ background and heavy augmentations.
\end{figure}

\begin{figure} [ht]
\includegraphics[height=55mm]{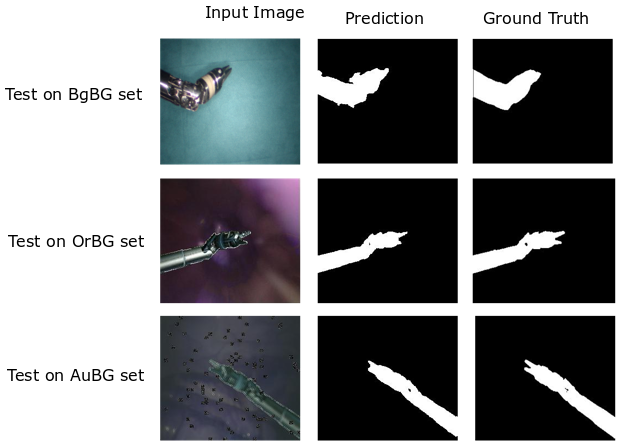}
\centering
\caption{U-Net segmentation overview}
\label{U-Net Segmentation Overview}
\footnotesize
Top: Test of segmentation on GrBG set.\break
Middle: Test of segmentation on OrBG set.\break
Bottom: Test of segmentation on AuBG set.\break
\end{figure}

On table \ref{tab:Unet_Simone_inigo_comparison} the averages of the best segmentation test results over the three test sets (GrBG, OrBG, and AuBG) are shown. A result is shown for each momentum $\beta$ used, with the batch size that yielded the best results. The choice of the momentum $\beta$ did not have as great an influence on the quality of the results as the choice of the batch size. Figure \ref{U-Net Segmentation Overview} shows a segmentation example of the U-Net performance on each test set. \\

\begin{table}[ht]
\small
\caption{Test results on Test datasets (GrBG, OrBG, and AuBG)}
\label{tab:Unet_Simone_inigo_comparison}
\centering
\begin{tabular}{l l}
Test & IoU Average \\
  \noalign{\global\arrayrulewidth=0.4mm}
\hline
Training 1: $\beta=0.5$, BZ = 8 & 88.03\% $\pm$ 11.27\%\\ 
  \noalign{\global\arrayrulewidth=0.1mm}
\hline
Training 2: $\beta=0.75$, BZ = 4 & 88.4\% $\pm$ 10.06\%\\ \hline
Training 3: $\beta=0.99$, BZ = 8 & 87.39\% $\pm$ 9.52\%\\ \hline

\end{tabular}
\end{table}

The results published by Hanisch \textit{et al.} \cite{hanischsegmentierung} offer a good benchmark to compare semantic segmentation results, since they performed semantic segmentation on the exact same MiroSurge surgical instrument. The authors were able to achieve a $77.76\% \pm 7.19\%$ IoU using a U-Net architecture trained on heavily augmented data. Averaging the segmentation results of our work yields an intersection over union of 87.94\%, which represents an improvement of 13.24\% on the results presented by Hanisch \textit{et al.} \cite{hanischsegmentierung}.

\subsection{DIFFERENT TOOL}

Result comparisons with other publications that perform semantic segmentation on surgical instruments can be performed to estimate the quality of the model w.r.t. state of the art approaches. Taking the datasets of the Robotic Instrument Segmentation subchallenge of the MiCCAi Conference, Pakhomov \textit{et al.} \cite{DBLP:journals/corr/PakhomovPAAN17} performed a cross validation test on the dataset, which is divided into six separate videos. Table shows the results of the multi-class semantic segmentation results.

\begin{table}[ht]
\small
\caption{Test results of Pakhomov \textit{et al.} \cite{DBLP:journals/corr/PakhomovPAAN17})}
\label{Test results of Pakhomov}
\centering
\begin{tabular}{l l l l l}

 & C1 & C2 & C3 & Mean \\
  \noalign{\global\arrayrulewidth=0.4mm}
\hline
Video 1 & 79.6\% & 68.2\% & 96.5\% & 81.4\%\\ 
  \noalign{\global\arrayrulewidth=0.1mm}
\hline
Video 2 & 82.2\% & 70.2\% & 98.6\% & 83.7\%\\  \hline
Video 3 & 80.4\% & 66.4\% & 98.0\% & 81.6\%\\  \hline
Video 4 & 75.0\% & 44.9\% & 97.1\% & 72.3\%\\  \hline
Video 5 & 72.3\% & 56.0\% & 93.6\% & 74.9\%\\  \hline
Video 6 & 70.7\% & 50.2\% & 95.7\% & 72.2\%\\  \hline
\end{tabular}
\end{table}

The authors thus achieve an IoU of 77.68\% over all classes and over all test videos.\\

\begin{figure} [ht]
\includegraphics[height=55mm]{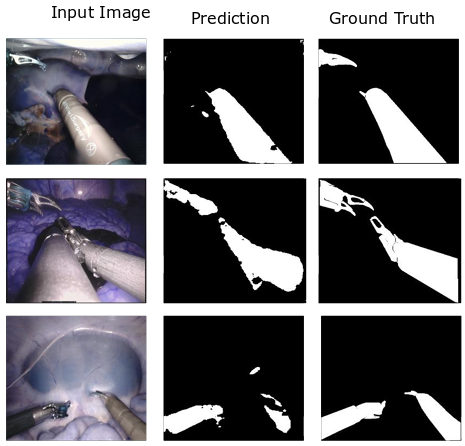}
\centering
\caption{Overview of the test results on da Vinci instruments}
\label{Test Results on da Vinci Instruments Overview}
\footnotesize

\end{figure}

By looking at tables \ref{tab:Unet_Simone_inigo_comparison} and \ref{Test results of Pakhomov} it can be seen that our model achieves state of the art precision on semantic segmentation. However, a direct comparison of the results is only possible if all the training assumptions are equal. In this case, the surgical instruments of the MediView system have different shape, texture, and color than the surgical instruments of the da Vinci system. The scope of our work is to perform binary, and not multi-class semantic segmentation as the authors of \cite{DBLP:journals/corr/PakhomovPAAN17} do. It is also not possible to quantitatively determine which surgical instruments are more challenging for a deep learning model to segment. Still, in order to get an overview and further test the generalization capabilities of our trained model, we tested it on the da Vinci dataset. The IoU results are summarized in table \ref{Test results on da Vinci Instruments}. As it was expected, the quality drops dramatically, due to the aforementioned differences on the surgical instruments. In several instances however, the model shows ability to detect most of the surgical instrument. Figure \ref{Test Results on da Vinci Instruments Overview} shows an overview of some of the best segmentation examples of the da Vinci surgical instruments. \\

\begin{table}[ht]
\small
\caption{Test results on da Vinci Instruments}
\label{Test results on da Vinci Instruments}
\centering
\begin{tabular}{l l}
Test & IoU Average\\
  \noalign{\global\arrayrulewidth=0.4mm}
\hline
Training 1: $\beta=0.5$, BZ = 8 & 49.05\% $\pm$ 8.97\%\\ 
  \noalign{\global\arrayrulewidth=0.1mm}
\hline
Training 2: $\beta=0.75$, BZ = 8 & 51.27\% $\pm$ 9.62\%\\ \hline
Training 3: $\beta=0.99$, BZ = 4 & 50.07\% $\pm$ 8.73\%\\ \hline

\end{tabular}
\end{table}

Despite the drop in semantic segmentation quality from an average 88.06\% IoU on the MediView instruments to an average 50.28 \% IoU on the da Vinici instruments, it is important to emphasize that a) the instruments of each system differ in shape, texture and color, and b) the training of our model was done using only one type of tool head, and one instrument per image, whereas the da Vinci dataset includes multiple instrument types, and one or more instruments per image. Given this circumstances, our model shows great potential at generalizing well to other previously unseen domains. Furthermore, it can be conjectured that a fine-tuning step with our model as base would greatly improve the results on any other type of surgical instruments. This step was however tested because of time limitations and because it fell outside the scope of this work, which was to test whether the approach presented would work on the MiroSurge surgical instruments.

\section{CONCLUSION AND FUTURE WORK}

In this work, a two step approach was proposed for 1) tackling semantic segmentation of surgical instruments, and 2) using domain adaptation techniques to reduce the overall cost of manually labeling training data and solving the problem of inaccurate mask labels of the dataset. \\

It was proved that it is possible to generate realistic-looking images from synthetic data to train a segmentation model that has state of the art generalization capabilities. To this end, a CycleGAN was used as the backbone of the domain adaptation step. To improve the results of the CycleGAN, an entire dataset of endoscopic and synthetic images was created, taking extra care that the distribution of the pose and geometry of the instruments on both datasets was as close as possible. A comprehensive dataset of generated images with high enough quality (realistic-looking appearance) was then created to be used as the training set for the semantic segmentation problem.\\

A U-Net model was then trained with the generated data, subject to heavy augmentations. The model was then tested on a set containing real endoscopic images of the surgical instruments, which yielded an 88.06\% IoU. This represents an improvement on some of the state-of-the-art results found in the literature, like \cite{hanischsegmentierung} and \cite{DBLP:journals/corr/PakhomovPAAN17}.\\

Additionally, the generalization capabilities of the segmentation model trained in the proposed fashion show potential even with images of completely different domains, such as the surgical instruments of the da Vinci surgical system. \\

The proposed method is thus best suited to tackle semantic segmentation problems where complex scenarios (like medical imaging) increase the time cost of manually labeling datasets, and in those cases where synthetic data of the system are available.\\

Several aspects can nevertheless be improved in future work.
During testing of the segmentation network we observed that the samples that yielded the lowest IoU were usually images with a larger than usual instrument scale. That is, the instrument was closer to the endoscope and thus occupied a much larger portion of the image. The MultiResUNet proposed by Ibtehaz \textit{et al.} \cite{ibtehaz2019multiresunet} could be worth investigating to solve the issue of scale invariance. On their work, they introduce an additional block of concatenated convolutions on the U-Net architecture ensuring that the features extracted at each layer come from different resolutions (or scales).\\

For the domain adaptation problem, a region proposal network \cite{DBLP:journals/corr/ZhouKLOT15} or attention maps \cite{lee2018davincigan} could be used to focus training of the model on the instrument, and avoid wasting computational power on analyzing the backgrounds.

\addtolength{\textheight}{-12cm}   








\newpage
\bibliography{example} 
\bibliographystyle{ieeetr}

\end{document}